\def\year{2021}\relax
\documentclass[letterpaper]{article} 
\usepackage{aaai21}  
\usepackage{times}  
\usepackage{helvet} 
\usepackage{courier}  
\usepackage[hyphens]{url}  
\usepackage{graphicx} 
\usepackage{natbib}  
\usepackage{caption} 
\usepackage{amssymb}
\usepackage{amsmath}
\usepackage{bm}
\usepackage{booktabs}
\usepackage{multirow}
\usepackage{tabularx}
\newcolumntype{R}{>{\raggedleft\arraybackslash}X}
\newcolumntype{C}{>{\centering\arraybackslash}X}
\usepackage{kotex}
\usepackage{color}

\usepackage{calc}
\newlength\bigH
\newlength\smlH
\title{Class-Attentive Diffusion Network for Semi-Supervised Classification}
\author{
    Jongin Lim\textsuperscript{\rm 1},
    Daeho Um\textsuperscript{\rm 1},
    Hyung Jin Chang\textsuperscript{\rm 2},
    Dae Ung Jo\textsuperscript{\rm 1},
    Jin Young Choi\textsuperscript{\rm 1}
    \\
}
\affiliations{
    \textsuperscript{\rm 1} Department of ECE, ASRI, Seoul National University \\
    \texttt{\{ljin0429,daehoum1,mardaewoon,jychoi\}@snu.ac.kr} \\
    \textsuperscript{\rm 2} School of Computer Science, University of Birmingham \\
    \texttt{h.j.chang@bham.ac.uk} \\
}

\begin{document}

\maketitle

\begin{abstract}

Recently, graph neural networks for semi-supervised classification have been widely studied.
However, existing methods only use the information of limited neighbors and do not deal with the inter-class connections in graphs.
In this paper, we propose Adaptive aggregation with Class-Attentive Diffusion (AdaCAD), a new aggregation scheme 
that adaptively aggregates nodes probably of the same class among K-hop neighbors.
%
To this end, we first propose a novel stochastic process, called Class-Attentive Diffusion (CAD), that strengthens attention to intra-class nodes and attenuates attention to inter-class nodes.
In contrast to the existing diffusion methods with a transition matrix determined solely by the graph structure, CAD considers both the node features and the graph structure with the design of our class-attentive transition matrix that utilizes a classifier.
Then, we further propose an adaptive update scheme that leverages different reflection ratios of the diffusion result for each node depending on the local class-context. 
As the main advantage, AdaCAD alleviates the problem of undesired mixing of inter-class features caused by discrepancies between node labels and the graph topology.
Built on AdaCAD, we construct a simple model called \textit{Class-Attentive Diffusion Network} (CAD-Net).
Extensive experiments on seven benchmark datasets consistently demonstrate the efficacy of the proposed method and our CAD-Net significantly outperforms the state-of-the-art methods.
Code is available at \url{https://github.com/ljin0429/CAD-Net}.

\end{abstract}

\section{Introduction}
Semi-supervised learning is a long-standing problem in machine learning.
Many semi-supervised learning algorithms rely on the geometry of the data induced by both labeled and unlabeled data points~\cite{chapelle2006semi}.
Since this geometry can be naturally represented by a graph whose nodes are data points and edges represent relations between data points, graph-based semi-supervised learning has been extensively studied for decades~\cite{zhu2003semi, zhou2004learning, belkin2006manifold, yang2016revisiting, kipf2016semi}.
In this paper, we focus on the problem of semi-supervised node classification on graphs. 
%

Graph Neural Networks (GNNs) have achieved remarkable progress in this field recently~\cite{ zhou2018graph, wu2020comprehensive, zhang2020deep}.
In particular, graph convolutions~\cite{kipf2016semi, gilmer2017neural} have received great attention due to its flexibility and good performances.
Underlying these methods is a neighborhood aggregation that forms a new representation of a node by aggregating features of itself and its neighbors.
The neighborhood aggregation is essentially a type of Laplacian smoothing~\cite{li2018deeper}, i.e., making the features of neighboring nodes similar, which makes the subsequent classification task easier. 
Built on this, several methods have developed the weighted aggregation using attention mechanisms~\cite{velivckovic2017graph, thekumparampil2018attention, zhang2018gaan} where the attention weights are determined by the features of each neighboring node pair.

However, one of the fundamental weaknesses with neighborhood aggregation methods is the lack of ability to capture long-range dependencies caused by over-smoothing~\cite{chen2019measuring}.
When stacking multiple layers to expand their range, the diameter of the smoothed area grows large, and eventually, the node representations in the entire graph become indistinguishable.
Although there have been miscellaneous efforts to overcome this issue~\cite{xu2018representation, abu2019mixhop, wu2019simplifying}, the range of these methods is still limited. 
Therefore, it is desirable to have the ability to propagate the label information over long-range, especially for large graphs or under sparsely labeled settings.

Recently, diffusion-based methods~\cite{klicpera2018predict, jiang2019data, klicpera2019diffusion} have demonstrated the capability of capturing long-range dependencies without leading to over-smoothing.
These methods utilize graph diffusion as an alternative to neighborhood aggregation.
Graph diffusion is a Markov process which spreads the information from the node to the adjacent nodes at each time step~\cite{masuda2017random}. 
Theoretically, $K$-steps of feature diffusion means that features of up to $K$-hop neighbors are aggregated to each node.
This aggregation-by-diffusion scheme allows the model to achieve a larger range without changing the neural network, whereas in the neighborhood aggregation scheme expanding the range would require additional layers. 
However, a major limitation of these methods is that they only utilize the graph structure with a transition matrix of diffusion determined by the graph adjacency matrix.
Since edges in real graphs are often noisy~\cite{khan2018uncertain} and could contain additional information,
there exist discrepancies between node labels and the graph structure~\cite{chen2019measuring}, i.e., some nodes may have more inter-class neighbors.
Thus, the aggregation scheme determined solely by the graph structure may lead to corrupted representations due to the inter-class connections.

To address the aforementioned limitation, we propose Class-Attentive Diffusion (CAD), a novel stochastic process that strengthens attention to intra-class nodes and attenuates attention to inter-class nodes by considering both the node features and the graph structure.
The proposed CAD attentively aggregates nodes probably of the same class among K-hop neighbors so that the feature representations of the same class become similar.
Then, we further propose a novel adaptive update scheme that assigns proper reflection ratios of the CAD result for each node depending on the local class-context.
If a node has many inter-class neighbors, our adaptation scheme puts more weights on the node's original feature than the aggregated feature by CAD and vice versa.
In this work, we call the overall scheme as \textit{\textbf{Ada}ptive aggregation with \textbf{CAD}} (AdaCAD).
Built on AdaCAD, we construct a simple model called \textit{Class-Attentive Diffusion Network} (CAD-Net).
Through extensive experiments, we validate the proposed method and show that AdaCAD enables the model to embed more favorable feature representations for better class separation.
Our CAD-Net significantly outperforms the state-of-the-art methods on 7 benchmark datasets from 3 different graph domains.

\section{Related Work}
\label{RelatedWork}
\subsection{Graph Neural Networks}
In recent literature on GNNs, there are two mainstreams: spectral-based methods and spatial-based methods.
The spectral-based methods~\cite{bruna2013spectral, henaff2015deep, defferrard2016convolutional, kipf2016semi} developed graph convolutions in the spectral domain using the graph Fourier transform.
However, these methods do not scale well with large graphs due to the computational burden.
The spatial-based methods~\cite{niepert2016learning, gilmer2017neural, hamilton2017inductive, velivckovic2017graph, monti2017geometric}, on the other hand, defined convolution-like operations directly on the graph based on the neighborhood aggregation.
The spatial-based methods, in particular, GCN~\cite{kipf2016semi}, MPNN~\cite{gilmer2017neural}, and SAGE~\cite{hamilton2017inductive} have received considerable attention due to its efficiency and superior performance.
Built on the neighborhood aggregation scheme, numerous variants have been proposed.
In the following, we categorize recent methods into three groups based on what they leverage to improve the model.

(\romannumeral 1) 
\textit{Extended Aggregation}.
Mixhop~\cite{abu2019mixhop} concatenates aggregated features from neighbors at different hops before each layer,
while in JK~\cite{xu2018representation}, skip connections are exploited to \textit{jump} knowledge to the last layer.
In SGC~\cite{wu2019simplifying}, multi-layers of GCN~\cite{kipf2016semi} are simplified into a single layer using the $K$-th power of an adjacency matrix, which means that the aggregation extends to the $K$-hop neighbor.
These methods use an extended neighborhood for aggregation.
However, the range of these methods is still limited, attributed to the low number of layers used.

(\romannumeral 2)
\textit{Feature Attention}.
Attention-based method such as GAT~\cite{velivckovic2017graph}, AGNN~\cite{thekumparampil2018attention}, and GaAN~\cite{zhang2018gaan}
have utilized attention mechanisms to develop weighted aggregation where the weighting coefficients are determined by the features of each neighboring node pair.
However, the aggregation of these methods is still limited to 1-hop neighbor. 
Meanwhile, Graph U-Nets~\cite{gao2019graph} proposed graph pooling and unpooling operations based on feature attention and then, developed an encoder-decoder architecture in analogy to U-Net~\cite{ronneberger2015u}.
However, the pooling operation proposed in their method does not take the graph structure into account but only depends on the node features~\cite{lee2019self}.

(\romannumeral 3)
\textit{Graph Diffusion}.
Recently, there have been several attempts utilizing graph diffusion.
These methods aggregate features by propagation over nodes using 
random walks~\cite{atwood2016diffusion, ying2018graph, ma2019pan}, 
Personalized PageRank (PPR)~\cite{klicpera2018predict}, 
Heat Kernel (HK)~\cite{xu2019graph}, 
and regularized Laplacian smoothing-based diffusion methods~\cite{jiang2019data}.
Meanwhile, GDC~\cite{klicpera2019diffusion} utilizes generalized graph diffusion (e.g. PPR and HK) to generate a new graph, then use this new graph instead of the original graph to improve performance.
However, all of the aforementioned methods do not take node features into account in their diffusion.

\subsection{Random Walks on Graph} 
Random walks have been extensively studied in classical graph learning; see~\cite{lovasz1993random, masuda2017random}
for an overview of existing methods.
In particular, random walks were used in the field of unsupervised node embedding~\cite{perozzi2014deepwalk, grover2016node2vec, tsitsulin2018verse, abu2018watch}. 
Unlike these methods, the proposed method aims to embed a more favorable node representation for semi-supervised classification. 
To achieve this, the proposed diffusion is class-attentive by considering the node features as well as the graph structure, while in those methods, it only depends on the graph adjacency matrix.
In~\cite{wu2012learning, wu2013analyzing}, Partially Absorbing Random Walk (PARW), a second-order Markov chain with partial absorption at each state, was proposed for semi-supervised learning. 
Co- \& Self-training~\cite{li2018deeper} utilized PARW for label propagation, presenting learning techniques that add highly confident predictions to the training set.
However, the state distribution of PARWs is also determined solely by the graph structure. 
Recently, several methods~\cite{lee2018graph, akujuobi2019collaborative, akujuobi2020recurrent} adopted reinforcement learning that aims to learn a policy that attentively selects the next node in the RW process. 
However, unlike the proposed method, their attention does not explicitly utilize class similarity since they employed additional modules to learn the policy.

\section{Proposed Method}
\subsection{Problem Setup}
Formally, the problem of semi-supervised node classification considers a graph $G = (\mathcal{V}, \mathcal{E}, X)$ where $\mathcal{V} = \{v_i\}_{i=1}^{N}$ is the set of $N$ nodes, $\mathcal{E}$ denotes the edges between nodes,
and $X \in \mathbb{R}^{N \times D}$ is a given feature matrix, i.e., $x_i$, $i$-th row of $X$, is $D$-dimensional feature vector of the node $v_i$.
Since edge attributes may not be given, we consider unweighted version of the graph represented by an adjacency matrix $A = [a_{ij}] \in \mathbb{R}^{N \times N}$ where $a_{ij} = 1$ if $\mathcal{E}(i,j) \neq 0$ and $a_{ij} = 0$ otherwise.
We denote the given label set as $Y_L$ associated with the labeled node set $\mathcal{V}_L$, i.e., $y_i \in Y_L$ be an one-hot vector indicating one of $C$ classes for $v_i$.
We focus on the transductive setting~\cite{yang2016revisiting} which aims to infer the labels of the remaining unlabeled nodes based on $(X, A, Y_L)$.

In general, the model for semi-supervised node classification can be expressed as
\begin{equation}
\label{eq:overview}
Z = f_{\theta}(X, A) \quad \text{and} \quad \hat{y}_i = g_\phi(z_i)\enskip (i = 1, 2, \cdots, N)
\end{equation}
where $f_{\theta}$ is a feature embedding network to embed the feature representations $Z$ from $(X, A)$, and $g_\phi$ is a node classifier predicting $\hat{y}_i$ from $z_i$, $i$-th row of $Z$.
The feature embedding network $f_{\theta}$ is realized by GNNs in recent literature.
The process of GNN can be decomposed into two steps: feature transformation and feature aggregation
where the former stands for a non-linear transformation of node features and the latter refers to the process of forming new representations via aggregating proximal node features.

In this paper, we focus on the process of feature aggregation.
More specifically, we aim to design an aggregation scheme which can be applied right before the classifier from Eq. (\ref{eq:overview}) to embed more favorable feature representations for class separation.
To this end, we first propose a novel Class-Attentive Diffusion (CAD), which attentively aggregates nodes probably of the same class among K-hop neighbors so that the representations of the same class become similar (see Section~\ref{sec:CAD}).  
Given $Z \in \mathbb{R}^{N \times F}$, CAD produces new feature representations $Z^{\mathrm{(CAD)}} \in \mathbb{R}^{N \times F}$ as follows,
\begin{equation}
\label{eq:defCAD}
Z^{\mathrm{(CAD)}} \leftarrow \mathrm{CAD}(Z, A, \{g_{\phi}(z_i)\}_{i=1}^{N}).
\end{equation}
Note that the node features ($Z$), the graph structure ($A$), and the class information ($g_{\phi}$) are jointly utilized.
Then, we further propose \textbf{Ada}ptive aggregation with \textbf{CAD} (AdaCAD) that leverages different reflection ratios of the diffusion result for each node depending on the local class-context (see Section~\ref{sec:AdaCAD}).
That is, AdaCAD produces the final feature representations $Z^{\mathrm{(AdaCAD)}} \in \mathbb{R}^{N \times F}$ as follows,
\begin{equation}
\label{eq:defAdaCAD}
Z^{\mathrm{(AdaCAD)}} \leftarrow \mathrm{AdaCAD}(Z, Z^{\mathrm{(CAD)}}, \Gamma)
\end{equation}
where $\Gamma$ assigns proper weights between $Z$ and $Z^{\mathrm{(CAD)}}$ for each node.
Built on AdaCAD, we construct a simple model called \textit{Class-Attentive Diffusion Network} in Section~\ref{sec:CAD-Net}.

\subsection{Class-Attentive Diffusion}
\label{sec:CAD}
In this section we present a novel stochastic process called Class-Attentive Diffusion (CAD), which combines the advantages of both the attention mechanism and the diffusion process.
The proposed CAD consists of $N$ Class-Attentive Random Walks (CARWs) starting at each node in the graph.
For clarity, we first explain how a single CARW is defined.

Suppose a CARW that starts from the node $v_i$.
The walker determines the next node among the neighbor by comparing the class likelihood given the node features, i.e., comparing $\mathbf{p}_i = p(y_i|z_i)$ and $\mathbf{p}_j = p(y_j|z_j)$ for $j \in \mathcal{N}(i)$. 
Our design objective is that the more similar $\mathbf{p}_i$ and $\mathbf{p}_j$, the more likely the walker moves from $v_i$ to $v_j$.
To this end, we define the transition probability from $v_i$ to $v_j$ as
\begin{equation}
\label{eq:transition}
T_{ij} = \mathrm{softmax}_{j \in \mathcal{N}(i)}(\mathbf{p}_i^T \mathbf{p}_j).
\end{equation}
Note that, $\mathbf{p}_i$ is a categorical distribution of which $c$-th element $\mathbf{p}_i(c)$ is the probability of node $i$ belongs to class $c$.
Thus, the cosine distance between $\mathbf{p}_i$ and $\mathbf{p}_j$ (i.e., $\mathbf{p}_i^T\mathbf{p}_j$) can be one possible solution for measuring the similarity between them. 
However, the true class likelihood $\mathbf{p}_i$ is intractable.
Instead, we approximate the true distribution by exploiting the classifier $g_{\phi}$ in Eq. (\ref{eq:overview})
where the probability of each class is inferred by $g_{\phi}$ based on the node feature $z_i$. 
That is,
\begin{equation}
\label{eq:self-guided}
\mathbf{p}_i \approx p(\hat{y}_i | z_i) = g_\phi(z_i).
\end{equation}
As the learning progresses, the transition matrix in Eq. (\ref{eq:transition}) gradually becomes more class-attentive by means of $g_{\phi}$.
This is the key difference from the recent diffusion-based methods, APPNP~\cite{klicpera2018predict}, GDEN~\cite{jiang2019data}, and GDC~\cite{klicpera2019diffusion},
where the transition matrix is determined solely by the adjacency matrix.

Let a row vector $\pi_{i}^{(t)} \in \mathbb{R}^{N}$ be the state distribution of the CARW after $t$ steps. 
This can be naturally derived by a Markov chain, i.e., $\pi_{i}^{(t+1)} = \pi_{i}^{(t)} T$,
where the initial state distribution $\pi_{i}^{(0)}$ be a one hot vector indicating the starting node $v_i$.
Then, this can be naturally extended to CAD
where $\bm{\Pi}^{(K)} \in \mathbb{R}^{N \times N}$ be the state distribution matrix after $K$-steps of CAD with entries $\bm{\Pi}^{(K)}(i,j) = \pi_{i}^{(K)}(j)$.

Now, we can define a new aggregation method with $K$-steps of CAD,
which forms a new feature representation of the node $v_i$ as follows,
\begin{equation} 
\label{eq:CAD}
z_i^{\mathrm{(CAD)}} = \sum_{j}\pi_{i}^{(K)}(j)\cdot z_j.
\end{equation}
Note that $\pi_{i}^{(K)}(j)$ is zero for $v_j$ beyond $K$-hop from $v_i$.
Hence, $\pi_{i}^{(K)}(j)$ naturally reflects the class similarity as it grows with the similarity between $\mathbf{p}_i$ and $\mathbf{p}_j$.
That is, $z_i^{\mathrm{(CAD)}}$ is essentially an attentive aggregation of $K$-hop neighbors where CAD strengthens attention to intra-class nodes and attenuates attention to inter-class nodes.

\subsection{Adaptive Aggregation with CAD}
\label{sec:AdaCAD}
In this section, we present Adaptive aggregation with CAD (AdaCAD).
We start by introducing our motivation.
In real graphs, some nodes may be connected to nodes of various classes, or even worse, nodes of the same class may not even exist in their neighbors.
Intuitively, in these cases, aggregated features from neighbors may lead to corrupted representations due to the inter-class connections.
Therefore, it should be needed to adaptively adjust the degree of aggregation for each node depending on its local class-context.

Motivated by this, we define AdaCAD to form a new feature representation of the node $v_i$ as follows,
\begin{equation} 
\label{eq:AdaCAD}
z_i^{\mathrm{(AdaCAD)}} = (1-\gamma_i) \cdot z_i + \gamma_i \cdot z_i^{\mathrm{(CAD)}}.
\end{equation}
Here, $\gamma_{i} \in [0, 1]$ controls the trade-off between its own node feature $z_i$ and the aggregated feature $z_i^{\mathrm{(CAD)}}$ from Eq. (\ref{eq:CAD}) by considering the local class-context of $v_i$.
For the node with neighbors of the same class, $\gamma_i$ should be a large value to accelerate proper smoothing.
In the opposite situation, $\gamma_i$ should be adjusted to a small value to preserve its original feature and avoid undesired smoothing.

To this end, we define a control variable $c_i$ as
\begin{equation}
c_i = \frac{1}{\mathrm{deg}(i)}\sum_{j \in \mathcal{N}(i)}g_{\phi}(z_i)^{T} g_{\phi}(z_j)
\end{equation}
where $\mathrm{deg}(i)$ is the degree of $v_i$ and $g_{\phi}$ is the aforementioned classifier.
%
Then, the range of $c_i$ would be $0 \leq c_i \leq 1$.
The meaning of $c_i$ is that the more nodes of the same class in the neighborhood, the greater the value of $c_i$ and vice versa. 
%
%
Therefore, we set up an adaptive formula for $\gamma_i$ as
\begin{equation}
\gamma_i = (1-\beta)c_i + \beta \gamma_{u}
\end{equation}
where $\gamma_u=1$ is the upper bound of $\gamma_i$ to keep $0 \leq \gamma_i \leq 1$ for interpolation of each node feature and the diffusion result.  
Note that $\gamma_i$ divides $c_i$ and $\gamma_u$ internally in the ratio of $\beta : (1-\beta)$
where $\beta \in [0, 1]$ controls the sensitivity of how much $\gamma_i$ will be adjusted according to $c_i$.
Since different graphs exhibit different neighborhood structures~\cite{klicpera2018predict}, the sensitivity $\beta$ is determined empirically for each dataset.

Now, we conclude the section with the overall formula of the proposed AdaCAD in a matrix form.
By letting $\bm{\Gamma} = \text{diag}(\gamma_1, \gamma_2, \cdots, \gamma_N)$ and combining Eq. (\ref{eq:CAD}) and (\ref{eq:AdaCAD}) together,
the entire aggregation scheme of AdaCAD can be expressed as follows,
\begin{equation}
\label{eq:matrixform}
Z^{\mathrm{(AdaCAD)}} = (\mathbf{I} - \bm{\Gamma})\cdot Z + \bm{\Gamma} \cdot \bm{\Pi}^{(K)} \cdot Z
\end{equation}
where $\bm{\Pi}^{(K)}$ is the state distribution matrix after $K$-steps of CAD.
Note that AdaCAD does not require additional learning parameters since we utilize the classifier $g_{\phi}$.

\subsection{Class-Attentive Diffusion Network}
\label{sec:CAD-Net}
Built on AdaCAD, we construct \textit{Class-Attentive Diffusion Network} (CAD-Net) for semi-supervised classification.
CAD-Net consists of the feature embedding network $f_{\theta}$ followed by AdaCAD and the classifier $g_{\phi}$ as defined in Eq. (\ref{eq:overview}), (\ref{eq:defAdaCAD}) and (\ref{eq:matrixform}).
More specifically, we realize $f_{\theta}$ with 2-layers of MLP for simplicity, as the process of feature aggregation can be sufficiently performed in AdaCAD,
and $g_{\phi}$ is realized by the softmax function, i.e., $\hat{y_i} = g_{\phi}(z_i) = \mathrm{softmax}(z_i)$ as in other literature~\cite{kipf2016semi, wu2019simplifying, jiang2019data}, and thus the dimension of $z_i$ is set to the number of classes.
The whole network parameters can then be trained in an end-to-end manner by minimizing the cross-entropy loss function $\mathcal{L}_{\mathrm{sup}}$ over all labeled nodes. 
By minimizing the cross-entropy between the label $y_i$
and the prediction $\hat{y_i}$ for all $y_i \in Y_L$, the model can be learned to enhance the element of $z_i$ that corresponds to the index indicating the class of $y_i$, which facilitates the class separation. 
In addition to $\mathcal{L}_{\mathrm{sup}}$, we consider another regularization objective.
As defined in Eq. (\ref{eq:transition}) and (\ref{eq:self-guided}), the transition matrix of CAD is determined by $\mathbf{p}_{i}$
where the initial distribution $\mathbf{p}_{i}$ for each node should generally be close to a one-hot vector such that the resulting transition matrix becomes more class-attentive.
Thus, we regularize the entropy of $\mathbf{p}_{i}$ by minimizing $\mathcal{L}_{\mathrm{ent}} = \sum_{i=1}^{N} H(\mathbf{p}_{i})$ where $H$ denotes the entropy function.
During training, $\mathcal{L}_{\mathrm{sup}}$ and $\mathcal{L}_{\mathrm{ent}}$ are jointly minimized by using Adam optimizer~\cite{kingma2014adam}.
We report the detailed implementation in Appendix A.\footnote{\url{https://github.com/ljin0429/CAD-Net}}

\section{Experiments}

\begin{table}
\caption{Dataset statistics.}
\label{dataset-table}
\centering
\footnotesize
\vspace{-1.5mm}
\begin{tabular}{lcccc} 
    \toprule
    Dataset                     &  Nodes      &   Edges     &   Features   &  Classes    \\
    \midrule
    {\tt\sc CiteSeer}           &  3327         &    4552       &    3703       &    6      \\
    {\tt\sc Cora}               &  2708         &    5278       &    1433       &    7      \\
    {\tt\sc PubMed}             &  19717        &    44324      &    500        &    3      \\
    {\tt\sc Amazon Comp.}       & 13752         &    245861     &    767        &    10     \\
    {\tt\sc Amazon Photo}       & 7650          &    119081     &    745        &    8      \\
    {\tt\sc Coauthor CS}        & 18333         &    81894      &    6805       &    15     \\
    {\tt\sc Coauthor Phy.}      & 34493         &    247962     &    8415       &    5      \\
    \bottomrule
\end{tabular}
\end{table}
\subsection{Datasets}
We conducted experiments on 7 benchmark datasets from 3 different graph domains: 
\textit{Citation Networks} (CiteSeer, Cora, and PubMed),
\textit{Recommendation Networks} (Amazon Computers and Amazon Photo), 
and \textit{Co-authorship Networks} (Coauthor CS and Coauthor Physics). 
\textbf{CiteSeer}, \textbf{Cora}, and \textbf{PubMed} are citation networks where each node represents a document and each edge represents a citation link.
Node features are bag-of-words descriptors of the documents, and class labels are given by the document's fields of study.
\textbf{Amazon Computers} and \textbf{Amazon Photo} are segments of Amazon co-purchase graph.
Here, each node represents a product and each edge indicates that two goods are frequently bought together.
Node features are bag-of-words descriptors which encode the product reviews, and class labels are given by the product category.
\textbf{Coauthor CS} and \textbf{Coauthor Physics} are co-authorship networks based on MS Academic Graph where each node represents an author and an edge is connected if they have co-authored a paper. 
Node features represent paper keywords for each author's papers, and class labels indicate the most active fields of study for each author.
Table~\ref{dataset-table} summarizes the dataset statistics.

\begin{figure*}[t]
\begin{center}
\includegraphics[width = 0.89\linewidth]{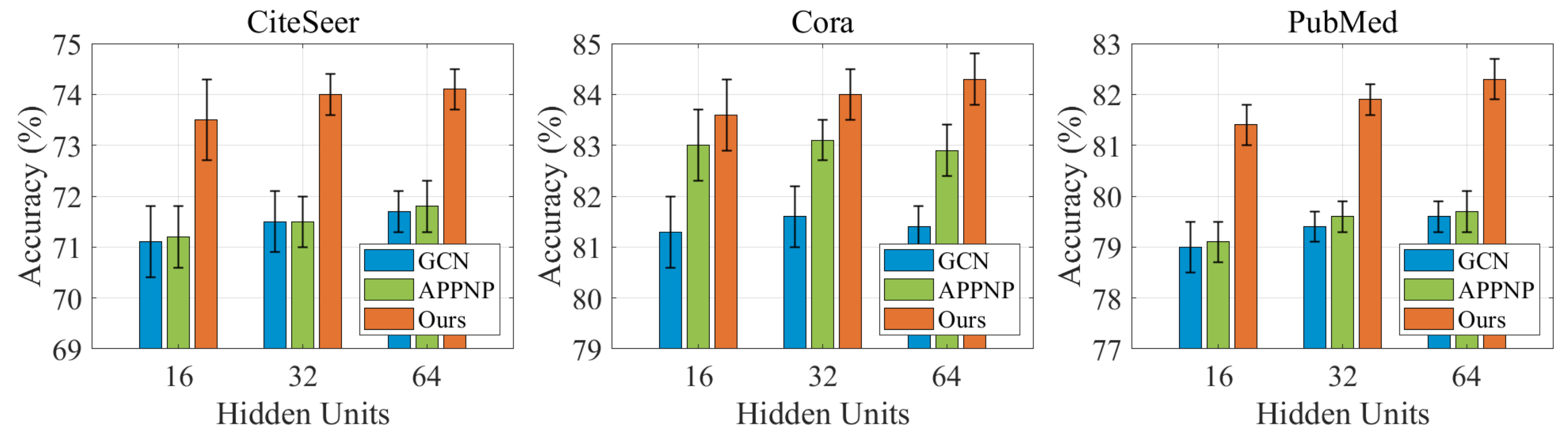}
\end{center}
\vspace{-4.0mm}
\caption{Accuracy (\%) with different hidden units. Note that GCN, APPNP, and the proposed CAD-Net have the same number of parameters. For all datasets, CAD-Net significantly outperform GCN and APPNP with the same number of parameters.} 
\label{fig:hidden}
\end{figure*}

\subsection{Experimental Setup}
For citation networks, we followed the standard benchmark setting suggested in~\cite{yang2016revisiting}.
We evaluated on the same train/validation/test split, which uses 20 nodes per class for train, 500 nodes for validation, and 1000 nodes for test.
For the credibility of the results, we report the average accuracy (\%) with the standard deviation evaluated on 100 independent runs.

For recommendation and co-authorship networks, we closely followed the experimental setup in~\cite{chen2019measuring}. 
We used 20 nodes per class for train, 30 nodes per class for validation, and the rest nodes for test.
We randomly split the nodes and report the average accuracy (\%) with the standard deviation evaluated on 100 random splits.

We compared the proposed method with the following 12 state-of-the-art methods:
\textbf{Cheby}~\cite{defferrard2016convolutional}, 
\textbf{GCN}~\cite{kipf2016semi}, 
\textbf{SAGE}~\cite{hamilton2017inductive},
\textbf{JK}~\cite{xu2018representation}, 
\textbf{MixHop}~\cite{abu2019mixhop}, 
\textbf{SGC}~\cite{wu2019simplifying},
\textbf{AGNN}~\cite{thekumparampil2018attention}, 
\textbf{GAT}~\cite{velivckovic2017graph}, 
\textbf{Graph U-Nets}~\cite{gao2019graph},
\textbf{APPNP}~\cite{klicpera2018predict}, 
\textbf{GDC}~\cite{klicpera2019diffusion}, 
and \textbf{GDEN}~\cite{jiang2019data}.
In all experiments, the publicly released codes were employed.

\subsection{Model Analysis}
In this section, we provide comprehensive analysis of the proposed method on CiteSeer, Cora, and PubMed
as they are the most widely used benchmark datasets in the literature.

\subsubsection{Influence of AdaCAD.}
To verify the effectiveness of AdaCAD, we compared AggCAD with 7 different aggregation methods. 
For a fair comparison, only AdaCAD is replaced with the same CAD-Net architecture.
Firstly, we consider 4 diffusion methods including Random Walks (RW), symmetric Normalized Adjacency matrix (symNA), Personalized PageRank (PPR)~\cite{page1999pagerank}, and Heat Kernel (HK)~\cite{kondor2002diffusion}.
For RW and symNA, the transition matrix is defined as $D^{-1}A$ and $D^{-\frac{1}{2}}AD^{-\frac{1}{2}}$ respectively, and we proceed $K$-steps of feature diffusion according to their transition matrix.
For PPR and HK, the closed-form solution of the diffusion state distribution is used as in~\cite{klicpera2019diffusion}.
Secondly, we consider 2 attentive diffusion variants. 
To the best of our knowledge, CAD is the first attempt that incorporates the feature attention and the diffusion process.
Therefore, we construct GAT+RW and TF+RW based on GAT~\cite{velivckovic2017graph} and Transformer~\cite{vaswani2017attention} respectively.
In GAT+RW, the transition is defined by the attention value computed by GAT, and we proceed $K$-steps of feature diffusion according to it.
Likewise, in TF+RW, the transition is defined by Transformer-style attention, i.e., $T_{ij} = \mathrm{softmax}_{j}(f_{\mathrm{Q}}(z_i)^T f_{\mathrm{K}}(z_j))$.
Lastly, we consider the model that only uses CAD for aggregation.

Table~\ref{diffusion-variants} shows the overall comparisons with the aforementioned variants.
Compared to RW, sym, PPR, and HK, which only utilize the graph structure, our variants (CAD-only and AggCAD) show superior results.
The better performance comes from the proposed class-attentive transition matrix both utilizing node features and the graph structure.
While GAT+RW and TF+RW can utilize both node features and the graph structure, the performances are not sufficient, which demonstrate the effectiveness of our design of class-attentive diffusion. 
Lastly, AdaCAD shows better performance than only using CAD.
By means of $\Gamma$ in AdaCAD, the model prevents undesired mixing from inter-class neighbors, which provides additional performance gains to CAD.

\begin{table}
\caption{Accuracy (\%) with different aggregation methods. 
Note that only the aggregation part (AdaCAD) is switched from the same CAD-Net architecture. 
}
\label{diffusion-variants}
\centering
\footnotesize
\vspace{-1mm}
\begin{tabular}{lccc} 
    \toprule
                        &  {\tt\sc CiteSeer}      &   {\tt\sc Cora}     &   {\tt\sc PubMed} \\
    \midrule
    RW      
        &   $71.5\pm0.5$ 
        &   $82.4\pm0.5$ 
        &   $79.6\pm0.4$       
        \\
    symNA   
        &   $71.4\pm0.6$ 
        &   $82.1\pm0.5$ 
        &   $79.8\pm0.3$       
        \\
    PPR     
        &   $72.5\pm0.9$ 
        &   $82.8\pm0.6$ 
        &   $79.3\pm0.6$        
        \\
    HK      
        &   $71.8\pm0.5$ 
        &   $82.3\pm0.6$ 
        &   $79.4\pm0.5$        
        \\
    \midrule
    GAT+RW  
        &   $70.2\pm1.4$ 
        &   $81.3\pm1.4$ 
        &   $77.7\pm1.0$        
        \\
    TF+RW   
        &   $70.4\pm1.5$ 
        &   $82.8\pm1.2$ 
        &   $78.6\pm1.1$       
        \\
    \midrule
    CAD     
        &   $73.5\pm0.5$ 
        &   $83.7\pm0.5$ 
        &   $80.2\pm0.5$       
        \\
    AdaCAD  
        &   $\mathbf{74.1\pm0.4}$ 
        &   $\mathbf{84.3\pm0.5}$ 
        &   $\mathbf{82.3\pm0.4}$   
        \\
    \bottomrule
\end{tabular}
\end{table}

\begin{figure*}[t]
\begin{center}
\includegraphics[width = 0.9\linewidth]{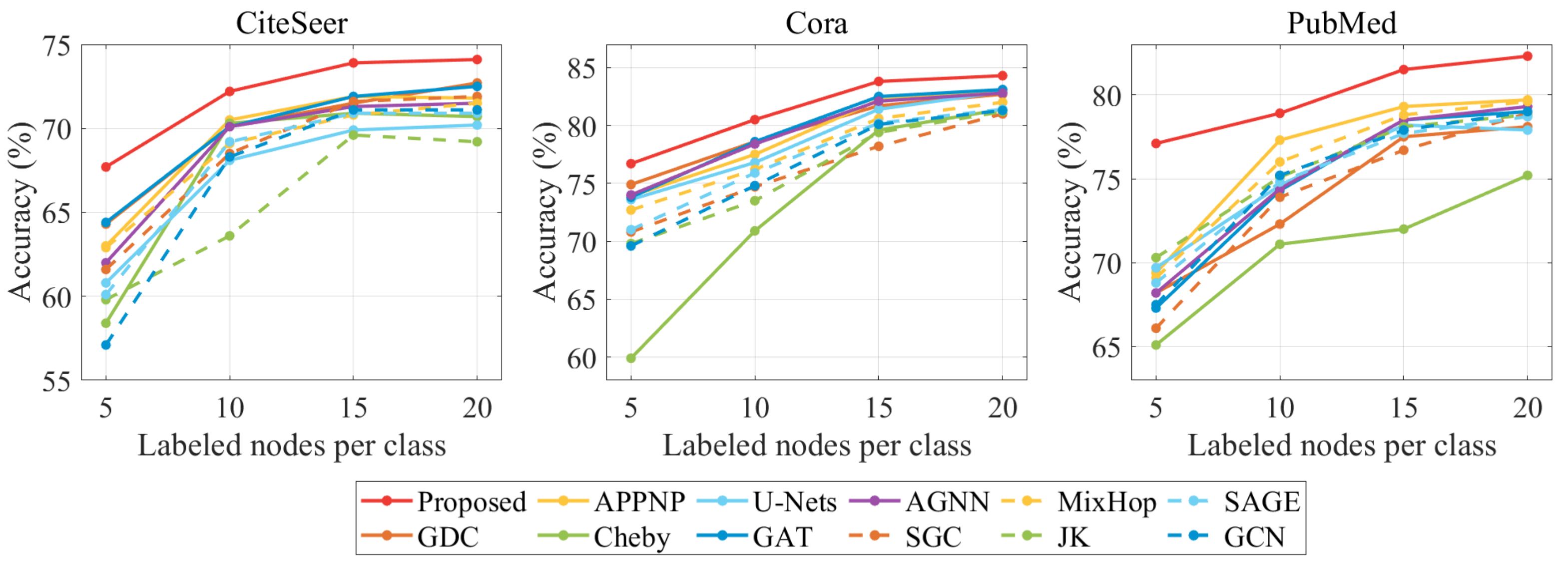}
\end{center}
\vspace{-3.0mm}
\caption{Accuracy (\%) with different numbers of labeled nodes per class. The proposed CAD-Net shows robust and superior performance for all settings.}
\label{fig:label}
\end{figure*}

\begin{figure*}
\begin{center}
\includegraphics[width = 0.89\linewidth]{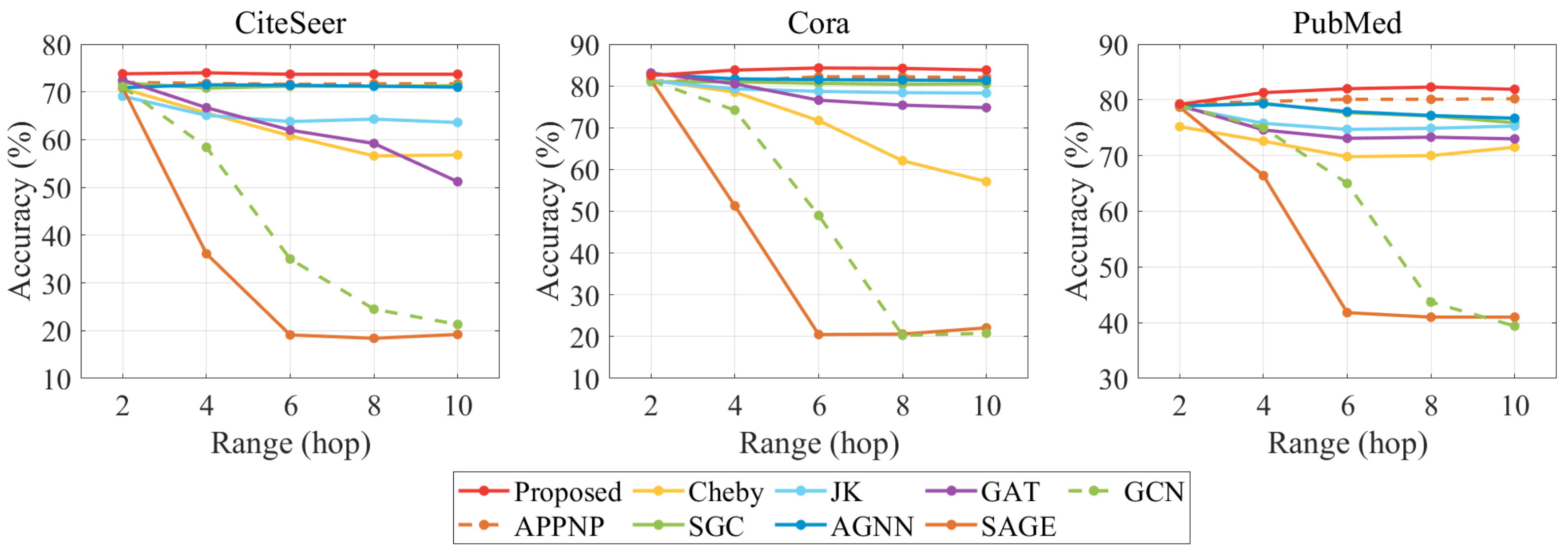}
\end{center}
\vspace{-3.0mm}
\caption{Accuracy (\%) with varying ranges of the model. The proposed CAD-Net shows superior performance regardless of the different ranges. The dominance of CAD-Net increases for a longer range.} 
\label{fig:range}
\end{figure*}

\subsubsection{Influence of Hidden Units.}
Unlike attention-based methods (AGNN and GAT), the proposed CAD can be self-guided by the classifier without the need for additional parameters for attention.
Thus, the total number of parameters can be implemented in the same way as the vanilla GCN.
To validate the effectiveness of AdaCAD, we evaluated the performance across the different numbers of hidden units in the feature embedding network $f_{\theta}$, and compared the results with GCN and APPNP which have the same number of parameters. 
As shown in Figure~\ref{fig:hidden}, CAD-Net shows robust performance with respect to the number of hidden units. 
Further, for all experiments, we can observe that CAD-Net significantly outperforms GCN and APPNP with the same number of parameters.
This demonstrates that the superior performance of CAD-Net is attributed to the proposed AdaCAD, not the power of the feature embedding network.

\subsubsection{Influence of $\bm{\beta}$.}
We also analyzed the influence of the hyperparameter $\beta$ which controls the sensitivity of how much $\gamma$ will be adjusted.
Due to the space limit, we attach the results to Appendix B.4.
While the optimum differs slightly for each dataset, we consistently found that any $\beta \in [0.65, 0.95]$ achieves the state-of-the-art performances.

\subsubsection{Different Label Rates.}
We then explored how the number of training nodes per class impacts the accuracy of the models.
The ability to maintain robust performance even under very sparsely labeled settings is important.
We compared the performances when the number of labeled nodes per class is changed to 20, 15, 10, and 5.
The overall results are presented in Figure~\ref{fig:label}.
CAD-Net shows robust and superior performance even under the very sparsely labeled setting and outperforms all other methods.
Note that, the diffusion-based methods (APPNP and GDC) do not show satisfactory results despite their wide range.
This is because these methods only utilize the graph structure.
In contrast, the proposed method aggregates nodes from a wide range and the importance of each node reflects both node features and the graph structure, which contributes to the superior performances of CAD-Net.
Especially, the superiority of CAD-Net is more obvious in PubMed which is a large dataset.
This further demonstrates the effectiveness of the proposed method.

\begin{table*}[t]
\caption{Accuracy (\%) under standard benchmark setting.
For all experiments, we report the performance evaluated over 100 independent runs.
OOM denotes out-of-memory.
(*We report the numbers taken from their paper since the code is not available.)
}
\vspace{-1.0mm}
\label{result-table}
\centering
\footnotesize
\begin{tabularx}{\textwidth}{cl*{7}{X}} 
    \toprule
    Type     & Method   &  {\tt\sc CiteSeer}  &  {\tt\sc Cora}  &  {\tt\sc PubMed}  &  {\tt\sc Amazon Comp.}  &  {\tt\sc Amazon Photo}  &  {\tt\sc Coauthor CS}  &   {\tt\sc Coauthor Physics}\\
    \midrule
    \multirow{3}{*}{Vanilla}  
    & Cheby   
        &   $70.7\pm0.5$     
        &   $81.4\pm0.5$       
        &   $75.2\pm1.4$ 
        &   $76.2\pm2.4$ 
        &   $85.9\pm2.3$ 
        &   \scriptsize{OOM}    
        &   \scriptsize{OOM}  
        \\
    & GCN     
        &   $71.1\pm0.7$ 
        &   $81.3\pm0.7$ 
        &   $79.0\pm0.5$ 
        &   $78.7\pm3.3$ 
        &   $88.9\pm1.9$ 
        &   $91.3\pm0.6$ 
        &   $93.3\pm0.8$ 
        \\
    & SAGE    
        &   $70.9\pm0.7$ 
        &   $81.4\pm0.7$ 
        &   $78.7\pm0.4$ 
        &   $78.9\pm2.1$ 
        &   $89.4\pm1.8$ 
        &   $91.6\pm0.6$ 
        &   $93.1\pm0.8$ 
        \\
    \midrule
    \multirow{3}{*}{\shortstack[1]{Extended \\ Aggregation}}  
    & JK      
        &   $69.1\pm1.1$ 
        &   $81.2\pm0.8$ 
        &   $78.7\pm0.5$ 
        &   $79.0\pm3.7$ 
        &   $89.1\pm2.0$ 
        &   $91.7\pm0.5$ 
        &   $93.2\pm0.9$ 
        \\
    & MixHop  
        &   $71.5\pm0.8$ 
        &   $82.0\pm1.0$ 
        &   $79.4\pm0.5$ 
        &   $79.5\pm2.8$ 
        &   $88.8\pm1.7$ 
        &   \scriptsize{OOM}    
        &   \scriptsize{OOM}   
        \\
    & SGC     
        &   $71.9\pm0.1$ 
        &   $81.0\pm0.2$ 
        &   $78.9\pm0.1$ 
        &   $81.5\pm1.8$ 
        &   $90.0\pm1.5$ 
        &   $91.2\pm0.6$ 
        &   $92.9\pm1.0$ 
        \\
    \midrule
    \multirow{3}{*}{\shortstack[1]{Feature\\ Attention}}  
    & AGNN    
        &   $71.5\pm0.7$ 
        &   $82.8\pm0.6$ 
        &   $79.3\pm0.8$ 
        &   $73.5\pm2.7$ 
        &   $88.0\pm3.4$ 
        &   $92.1\pm0.6$ 
        &   $93.8\pm0.7$ 
        \\
    & GAT     
        &   $72.5\pm0.8$ 
        &   $83.1\pm0.8$ 
        &   $79.0\pm0.3$ 
        &   $80.5\pm2.2$ 
        &   $90.6\pm1.2$ 
        &   $91.0\pm0.5$ 
        &   $93.1\pm0.6$ 
        \\
    & U-Nets  
        &   $70.2\pm1.0$ 
        &   $82.9\pm0.7$ 
        &   $78.0\pm0.5$ 
        &   $76.9\pm2.1$ 
        &   $87.1\pm1.8$ 
        &   $91.7\pm1.0$ 
        &   $93.4\pm0.7$ 
        \\
    \midrule
    \multirow{3}{*}{\shortstack[1]{Graph\\ Diffusion}}  
    & APPNP   
        &   $71.8\pm0.5$ 
        &   $82.9\pm0.5$ 
        &   $79.7\pm0.3$ 
        &   $81.0\pm1.9$ 
        &   $90.5\pm1.6$ 
        &   $92.3\pm0.4$ 
        &   $93.5\pm0.7$ 
        \\
    & GDC     
        &   $72.7\pm0.8$ 
        &   $82.7\pm0.7$ 
        &   $78.1\pm0.3$ 
        &   $81.6\pm2.8$ 
        &   $88.8\pm1.7$ 
        &   \scriptsize{OOM}    
        &   $92.7\pm0.8$ 
        \\
    & *GDEN   
        &   $72.8$          
        &   $82.0$            
        &   $78.7$            
        &   -                  
        &   -            
        &   -                   
        &   -                 
        \\
    \midrule
    Proposed & CAD-Net 
        &   $\mathbf{74.1\pm0.4}$ 
        &   $\mathbf{84.3\pm0.5}$ 
        &   $\mathbf{82.3\pm0.4}$ 
        &   $\mathbf{82.1\pm2.0}$ 
        &   $\mathbf{90.9\pm1.5}$ 
        &   $\mathbf{93.5\pm0.6}$ 
        &   $\mathbf{94.7\pm0.4}$ 
        \\
    \bottomrule
\end{tabularx}
\end{table*}

\subsubsection{Different Ranges.}
Figure~\ref{fig:range} shows influence of the different ranges.
As expected, the neighborhood aggregation methods degrade performance with increasing layers.
While the diffusion-based methods maintain the performance with increasing ranges, CAD-Net shows superior performance for all ranges.
Also, as in the previous experiment, the superiority of CAD-Net is particularly evident in PubMed, which suggests that the proposed method is able to accommodate larger graphs or sparsely labeled settings.
%

\begin{table}[t]
\caption{Average accuracy (\%) evaluated on 100 \textit{Random} train/validation/test splits.
}
\label{random-splits}
\centering
\footnotesize
\vspace{-1.0mm}
\begin{tabular}{lccc} 
    \toprule
    Method            
        &  {\tt\sc CiteSeer}      
        &   {\tt\sc Cora}     
        &   {\tt\sc PubMed} 
        \\
    \midrule
    Cheby   
        & $69.1\pm1.9$         
        & $77.8\pm2.3$         
        & $73.1\pm3.2$   
        \\
    GCN
        & $68.3\pm1.9$ 
        & $79.3\pm1.8$ 
        & $77.2\pm2.6$ 
        \\
    SAGE
        & $68.8\pm1.8$ 
        & $79.8\pm1.7$  
        & $77.2\pm2.5$  
        \\
    \midrule
    JK  
        & $67.5\pm1.9$ 
        & $78.6\pm1.9$ 
        & $77.7\pm2.7$   
        \\
    MixHop
        & $68.4\pm1.7$  
        & $80.7\pm1.7$ 
        & $77.8\pm2.5$  
        \\
    SGC
        & $69.2\pm1.7$ 
        & $79.9\pm1.8$ 
        & $77.0\pm2.6$ 
        \\
    \midrule
    AGNN
        & $69.4\pm1.8$ 
        & $80.8\pm1.8$  
        & $78.0\pm2.4$  
        \\
    GAT   
        & $70.0\pm1.9$ 
        & $81.6\pm1.5$   
        & $77.3\pm2.4$ 
        \\
    U-Nets
        & $67.9\pm1.9$ 
        & $81.2\pm1.9$ 
        & $77.8\pm2.6$   
        \\
    \midrule
    APPNP 
        & $69.9\pm1.7$ 
        & $81.8\pm1.5$   
        & $78.8\pm2.5$ 
        \\
    GDC
        & $70.9\pm1.7$ 
        & $81.9\pm1.5$  
        & $76.9\pm2.4$  
        \\
    \midrule
    CAD-Net 
        & $\mathbf{71.1\pm1.6}$ 
        & $\mathbf{82.4\pm1.4}$  
        & $\mathbf{79.6\pm2.4}$  
        \\
    \bottomrule
\end{tabular}
\end{table}

\subsection{Comparison with State-of-the-art Methods}

\subsubsection{Evaluation on Benchmark Datasets.}
Table~\ref{result-table} shows the overall results under standard benchmark settings.
In all experiments, the proposed CAD-Net shows superior performance to other methods.
We also provide statistical analysis of the results in Appendix B.6, demonstrating that CAD-Net achieves statistically significant improvements. 
The better performance of CAD-Net comes from the proposed adaptive aggregation scheme based on the class-attentive diffusion both utilizing node features and the graph structure in the transition matrix.
In addition, we provide further comparisons with the latest methods~\cite{liu2020towards, chen2020simple, hassani2020contrastive, zhu2020bilinear, zhang2020adaptive} and our CAD-Net still achieves state-of-the-art performance (see Appendix B.7).

\subsubsection{Computational Complexity.}
In terms of memory requirement, CAD-Net is as efficient as APPNP with the same number of parameters (see Figure~\ref{fig:hidden}).
Only one forward operation is additionally required to obtain our class-attentive transition probability.
To further validate the computational efficiency of CAD-Net, we compared the average training time per epoch (ms) measured on a single Nvidia GTX 1080 Ti machine.  
As expected, we confirmed that CAD-Net is on par with APPNP and much faster than GAT.
The detailed results are provided in Appendix B.8.

\subsubsection{Random Splits.}
%
Recently, \cite{shchur2018pitfalls} pointed out that the data split (train, validation, test) has a significant influence on the performance.
Therefore, we further evaluated average accuracy computed over 100 \textit{Random} splits where the splits are \textit{randomly} drawn with 20 nodes per class for train, 500 nodes for validation, and 1000 nodes for test. 
As shown in Table~\ref{random-splits}, CAD-Net shows robust and superior performance regardless of the data splits.

\section{Conclusion}
In this paper, we propose Adaptive aggregation with Class-Attentive Diffusion (AdaCAD), a new aggregation scheme for semi-supervised classification on graphs.
The main benefits of the proposed AdaCAD are three aspects.
(\romannumeral 1) 
AdaCAD attentively aggregates nodes probably of the same class among $K$-hop neighbors employing a novel Class-Attentive Diffusion (CAD).
Unlike the existing diffusion methods, both the node features and the graph structure are leveraged in CAD with the design of the class-attentive transition matrix which utilizes the classifier.
(\romannumeral 2) 
For each node, AdaCAD adjusts the reflection ratio of the diffusion result differently depending on the local class-context, which prevents undesired mixing from inter-class neighbors.
(\romannumeral 3) 
AdaCAD is computationally efficient and does not require additional learning parameters since the class-attentive transition probability is defined by the classifier.
Extensive experimental results demonstrate the validity of AdaCAD and Class-Attentive Diffusion Network (CAD-Net), our simple model based on AdaCAD, achieves state-of-the-art performances by a large margin on seven benchmark datasets.

\section*{Acknowledgment}
This research was supported by the IITP (Institute for Information \& Communication Technology Promotion) grant funded by the MSIT (Ministry of Science and ICT, Korea): [2017-0-00306, Outdoor Surveillance Robots] and [IITP-2020-2020-0-01789, ITRC(Information Technology Research Center) support program].

\section*{Ethics Statement}
Graphs accommodate many potential real-world applications such as social networks and web pages.
Our research is a study of neural networks applicable in the graph domain.
Therefore, our research can be an important basis for graph-based applications to be applied in real life in the future.
Besides, a large amount of cost is required to acquire high quality of labeled data.
The problem of semi-supervised learning, which we focus on, can secure robust performance with a small number of labeled data, thus contributing to lowering the threshold of solving industrial or social problems using machine learning at a low cost.

\bibliography{refs}

\clearpage
\appendix
\onecolumn

\section{Implementation}
In this section, we present the detailed implementation of CAD-Net.
We used PyTorch Geometric~\cite{Fey/Lenssen/2019} for implementation.
Table~\ref{implementation} summarizes a full list of hyperparameters of CAD-Net used in the experiments.
For all datasets, 2-layers of MLP with the number of hidden units of 64 was used for $f_{\theta}$.
We used the leaky ReLU activation and dropout for the whole network.
In Table~\ref{implementation}, \textit{leak-slope} denotes the negative slope of the leaky ReLU and $p_{\rm{drop}}$ denotes the drop probability in dropout. 
The number of diffusion steps $K$ and the sensitivity $\beta$ were determined empirically.
Note that, unlike other methods, the proposed method explicitly interpolates the node's original feature (see Eq. (\ref{eq:AdaCAD}) of the manuscript) so that the self-loop is not necessary.
We used the graphs with self-loops for CiteSeer, Cora, and Amazon Photo datasets because their graphs contain a number of isolated regions where only few nodes are connected.
For training, we used Adam optimizer~\cite{kingma2014adam} with full-batch.
During training, we dropped the learning rate (lr) to 1/2 of its previous value every 100 epochs (CiteSeer and PubMed) and 50 epochs (Cora).
For a fair comparison to other methods, we did not use the lr drop strategy for Amazon and Coauthor datasets.
We used $L_2$ regularization on the learnable parameters (weight decay).
In addition, $\lambda_{\mathrm{ent}}$ denotes the weighting parameter for our entropy regularization loss term $\mathcal{L}_{\mathrm{ent}}$.
As in other works~\cite{kipf2016semi, wu2019simplifying, klicpera2019diffusion}, 
we used early stopping with a window size reported in Table~\ref{implementation}.
If the early stop cell is empty, we trained the model without early stopping. 
%
%
%
Our code is available at \url{https://github.com/ljin0429/CAD-Net}.

\begin{table}[h]
\caption{Hyperparameters for CAD-Net used in the experiments.}
\label{implementation}
\centering
\footnotesize
\vspace{-1.5mm}
\begin{tabularx}{\textwidth}{l*{7}{C}} 
\toprule
    & {\tt\sc CiteSeer}            
    & {\tt\sc Cora}              
    & {\tt\sc PubMed}
    & {\tt\sc Amazon Comp.}
    & {\tt\sc Amazon Photo} 
    & {\tt\sc Coauthor CS}  
    & {\tt\sc Coauthor Physics} 
    \\
\midrule
layers                 
    & 2                            
    & 2                          
    & 2
    & 2
    & 2                          
    & 2
    & 2
    \\
hidden units           
    & 64                           
    & 64                         
    & 64
    & 64
    & 64
    & 64
    & 64
    \\
leak-slope            
    & 0.05                         
    & 0.05                       
    & 0.1   
    & 0.01
    & 0.15
    & 0.01
    & 0.01
    \\
$p_{\rm{drop}}$          
    & 0.3                          
    & 0.5                        
    & 0.3            
    & 0.3
    & 0.5
    & 0.3
    & 0.3
    \\
$K$                      
    & 3                            
    & 6                          
    & 8        
    & 2
    & 2
    & 4
    & 6
    \\
$\beta$                   
    & 0.7                          
    & 0.8                        
    & 0.85     
    & 0.95
    & 0.95
    & 0.8
    & 0.8
    \\
self-loop              
    & o                            
    & o                          
    & x        
    & x
    & o
    & x
    & x
    \\
epochs                  
    & 200                          
    & 100                        
    & 300         
    & 300
    & 300
    & 100
    & 200
    \\
initial lr     
    & 0.03                         
    & 0.01                       
    & 0.03               
    & 0.03
    & 0.05
    & 0.02
    & 0.02
    \\
lr drop          
    & 100/0.5                         
    & 50/0.5                        
    & 100/0.5     
    & -
    & -
    & -
    & -
    \\
weight decay           
    & 5e-4                       
    & 5e-4                     
    & 5e-4      
    & 1e-5
    & 2e-7
    & 1e-6
    & 1e-6
    \\
$\lambda_{\mathrm{ent}}$ 
    & 0.3                          
    & 0.5                        
    & 0.5           
    & 0.1
    & 0.1
    & 0.7
    & 0.7
    \\
early stop           
    & -                            
    & 10                         
    & 30           
    & -
    & 20
    & 20
    & 20
    \\
\bottomrule
\end{tabularx}
\end{table}

\section{Experiments}

In this section, we provide further details of the experimental setup that were not presented in the manuscript due to the space limit. 
Moreover, we also present additional experiments and investigations on what happens inside AdaCAD.

\subsection{Datasets}
All datasets used in the experiments are included in PyTorch Geometric~\cite{Fey/Lenssen/2019}. 
Following PyTorch Geometric, we pre-processed the datasets by removing self-loops and duplicated edges from the graphs and used these pre-processed versions of the graphs for all experiments. 
We reported the dataset statistics in Table~\ref{dataset-table} of the manuscript.

\subsection{Learning Process}
\begin{figure}[ht]
\begin{center}
\includegraphics[width=0.37\textwidth]{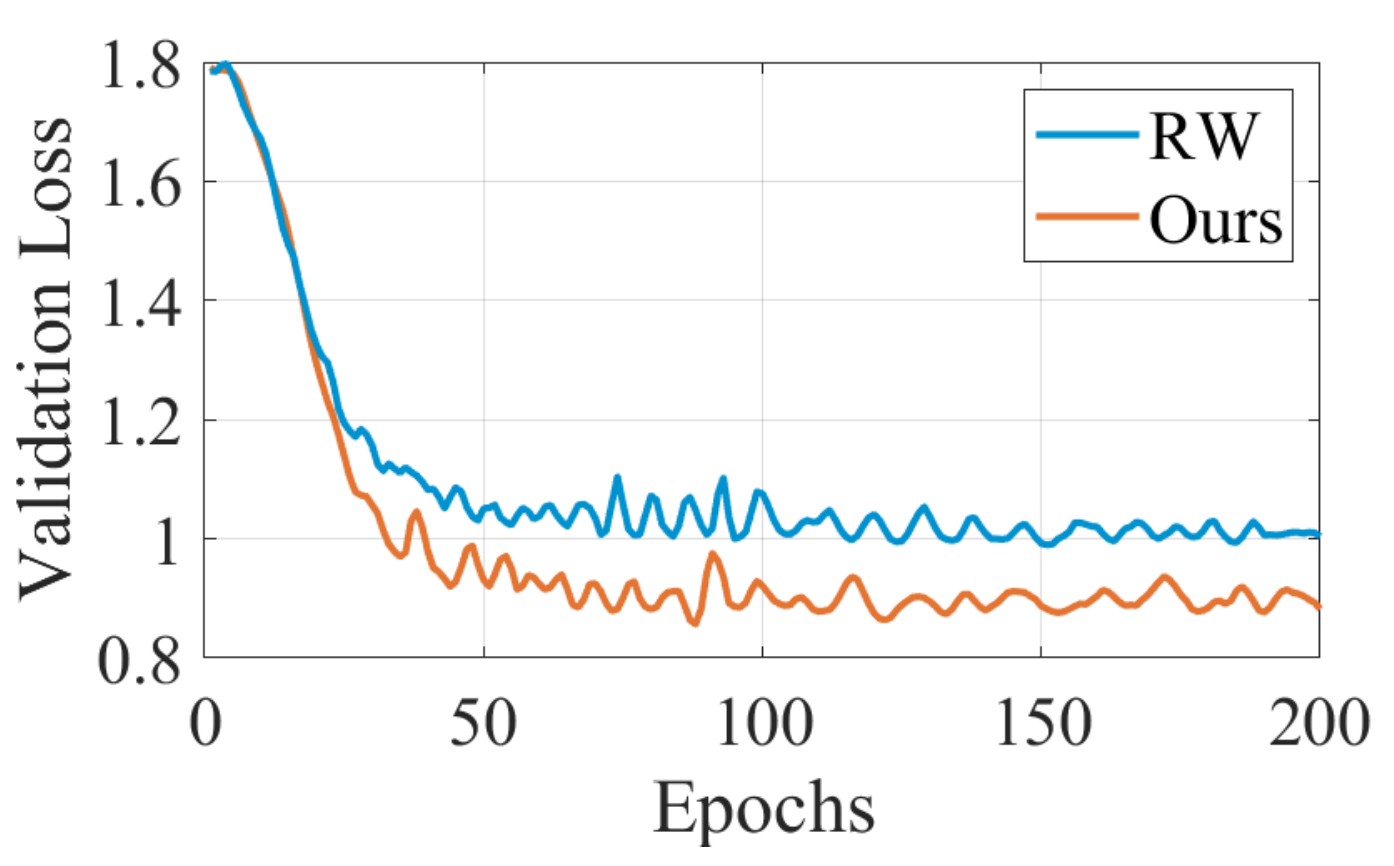}
\end{center}
\vspace{-3.5mm}
\caption{
Validation loss on {\tt\sc CiteSeer}.
Since the classifier is not trained at the beginning, there is no big difference from the vanilla RW for the first few epochs.
As the learning progresses, the class prediction becomes more accurate, and thus, the transition probability between the two nodes of the same class will become larger, i.e., the transition gradually becomes more class-attentive.
Consequently, our AdaCAD obtains obviously lower cross-entropy loss than vanilla RW model at convergence.
}
\label{fig:loss}
\end{figure}

Figure~\ref{fig:loss} shows the cross-entropy loss on validation set of CiteSeer across different epochs. 
We compare the learning curve of CAD-Net with RW model described in Section 4.3 of the manuscript.
Since the classifier is not trained at the beginning, the transition is almost uniformly random to all neighboring nodes. 
Thus, there is no difference from RW at the beginning.
As the classifier is trained by the given labeled nodes, the class prediction becomes more accurate.
Therefore, the transition of CAD becomes gradually class-attentive as the learning progresses, which leads to better predictive accuracy of CAD-Net.
As shown in Figure~\ref{fig:loss}, CAD-Net obtains obviously lower cross-entropy loss values at convergence.

\subsection{Entropy Regularization}
The proposed CAD-Net is trained to jointly minimize the cross-entropy loss $\mathcal{L}_{\mathrm{sup}}$ for given labeled nodes and entropy regularization loss $\mathcal{L}_{\mathrm{ent}}$ (see Section 3.4 of the manuscript).
In this section, we analyzed the effect of the entropy regularization term $\mathcal{L}_{\mathrm{ent}}$.
For comparision, we trained CAD-Net only with the cross-entropy loss $\mathcal{L}_{\mathrm{sup}}$, and then  evaluated the performance. 
We confirmed that the average accuracies (with std) are 72.1\% (0.6), 83.6\% (0.4), and 81.7\% (0.3), for CiteSeer, Cora, and PubMed, respectively. 
Without the entropy regularization, there is slight performance degradation since the high entropy of $\mathbf{p}_{i}$ gives a less class-attentive transition matrix.

\subsection{Influence of $\bm{\beta}$}
\begin{figure*}[h]
\begin{center}
\includegraphics[width = 0.91\linewidth]{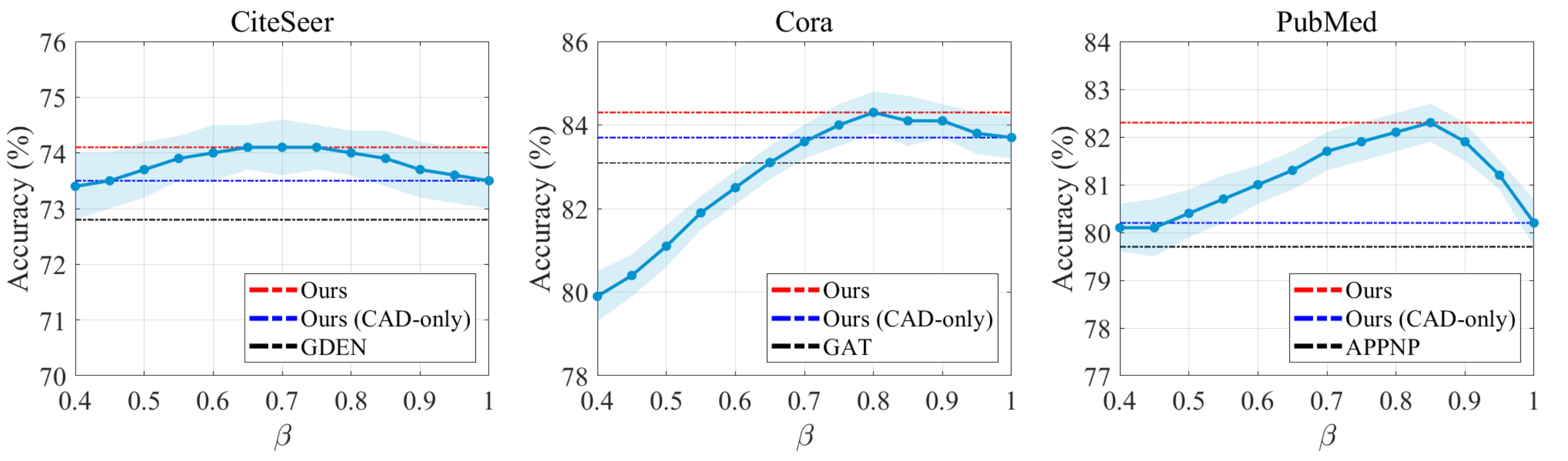}
\end{center}
\vspace{-4mm}
\caption{Accuracy with different $\beta$. For all datasets, $\beta \in [0.65, 0.95]$ consistently achieves the state-of-the-art performances.} 
\label{fig:beta}
\end{figure*}

\begin{figure*}[h]
\begin{center}
\includegraphics[width =0.9 \linewidth]{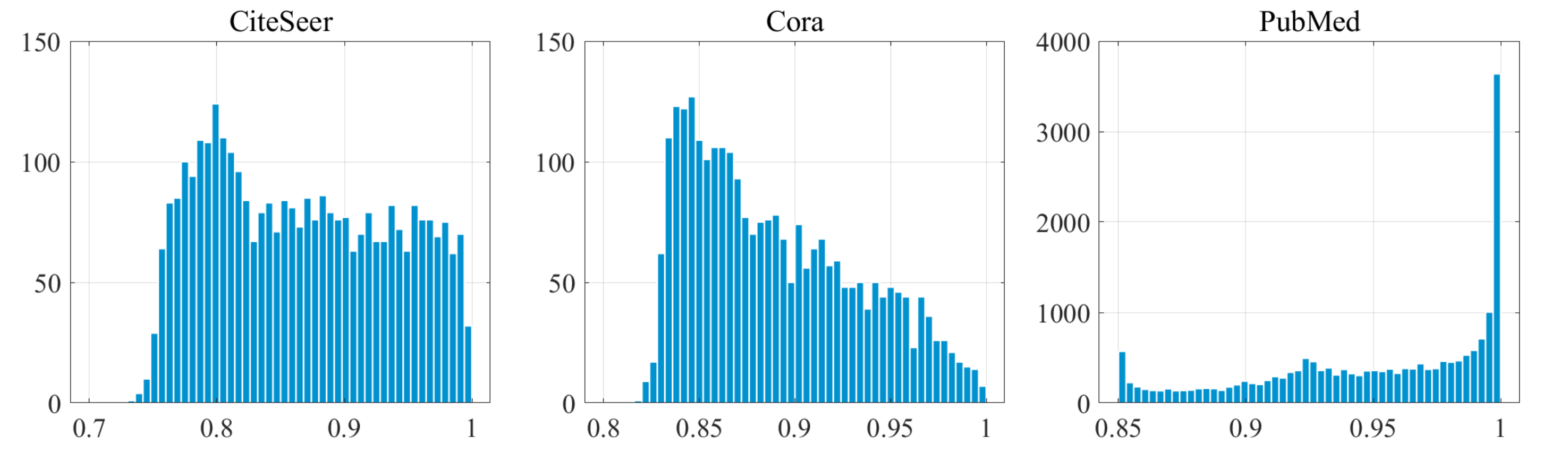}
\end{center}
\vspace{-3mm}
\caption{Histogram of $\gamma$ values for each dataset. Different datasets exhibit different distributions.}
\label{fig:histogram}
\end{figure*}

Figure~\ref{fig:beta} shows the effect of the hyperparameter $\beta$ which controls the sensitivity of how much $\gamma$ will be adjusted.
The blue line indicates the mean accuracy with different $\beta$ and the shaded area indicates the standard deviation computed over 100 independent runs.
The black dashed line indicates the existing state-of-the-art performance.
While the optimum value for $\beta$ differs slightly for each dataset, we can observe that any $\beta \in [0.65, 0.95]$ achieve the state-of-the-art performances.
Note that setting $\beta=1$ leads to $\gamma_i = 1$, which corresponds to CAD-only (blue dashed line) that only uses CAD without the adaptive aggregation in AdaCAD. 
The CAD-only model achieves state-of-the-art performance for all datasets, demonstrating that the proposed CAD is effective.
We further confirm that a proper $\beta$ leads to additional performance gain (red dashed line) for all datasets.
That is, by means of $\gamma$, the model can adjust the reflection ratio of the diffusion to prevent corrupted representations from inter-class neighbors, which leads to better performance.
Figure~\ref{fig:histogram} illustrates the distribution of the resulting $\gamma$ for entire nodes where $\gamma_i$ for each node varies reflecting the node's local class context.

\vspace{-1mm}

\pagebreak

\subsection{Influence of Incorrect Predictions}
The proposed CAD-Net utilizes the classifier's output to define a class-attentive transition matrix.
One would concern that CAD-Net might have a tendency to reinforce early incorrect decisions.
If the classifier makes an incorrect decision in the first place, the transition probability is poorly defined and can eventually prejudice the classification results.
However, the proposed method is hardly degenerated by the wrong predictions.
In a situation that the classifier makes an incorrect decision, the resulting prediction mostly has ambiguous likelihood distribution, i.e., $p_i$ has high entropy.
In such cases, the transition probability becomes low, reducing the attention to that node.
Therefore, our method is generally not overconfident with wrong predictions, which prevents the reinforcement of incorrect decisions.

\vspace{+1mm}
\subsection{Statistical Analysis}
For all experiments in the manuscript, we reported the average accuracy (\%) with the standard deviation evaluated on 100 independent runs for the credibility of the results.
To further demonstrate that the performance improvements of CAD-Net are statistically significant, we first investigated $p$-values of paired $t$-tests compared with the second-best model.
Besides, to ensure the statistical robustness of our experimental results, we calculated confidence intervals via bootstrapping. 
More specifically, we compared 95\% Bootstrap Confidence Intervals for the reported accuracies of the second-best model (BCI-2b) and those of ours (BCI-ours). 
Table~\ref{stats_analysis} summarizes the results, demonstrating that CAD-Net achieves statistically robust and significant improvements for all experiments.

\begin{table*}[h]
\caption{$p$-values of the paired $t$-test and 95\% bootstrap confidence intervals for the reported accuracies of the second-best model (BCI-2b) and those of ours (BCI-ours).}
\label{stats_analysis}
\centering
\footnotesize
\vspace{-1.0mm}
\begin{tabularx}{\textwidth}{l*{7}{C}} 
\toprule
    & {\tt\sc CiteSeer}            
    & {\tt\sc Cora}              
    & {\tt\sc PubMed}
    & {\tt\sc Amazon Comp.}
    & {\tt\sc Amazon Photo} 
    & {\tt\sc Coauthor CS}  
    & {\tt\sc Coauthor Physics} 
    \\
\midrule
$p$-value                 
    & 2.61e-34     
    & 2.78e-36                          
    & 1.04e-67
    & 0.03
    & 2.42e-05                          
    & 5.60e-39
    & 2.71e-32
    \\
BCI-2b            
    & (72.41, 72.70)                           
    & (82.74, 82.96)                         
    & (79.76, 79.92)
    & (81.37, 82.03)
    & (90.20, 90.77)
    & (92.19, 92.36)
    & (93.41, 93.67)
    \\
BCI-ours         
    & (74.07, 74.26)                         
    & (84.16, 84.35)                       
    & (82.19, 82.32)   
    & (81.73, 82.37)
    & (90.61, 91.20)
    & (93.35, 93.56)
    & (94.57, 94.73)
    \\
\bottomrule
\end{tabularx}
\end{table*}

\subsection{Comparison with 2020 Methods}
As requested by the reviewer, we further compared with the following five major methods published in 2020 that share the same experimental settings as ours;

\begin{itemize}
\item DAGNN~\cite{liu2020towards}: \textit{Towards Deeper Graph Neural Networks} [KDD’20]
\item GCNII~\cite{chen2020simple}: \textit{Simple and Deep Graph Convolutional Networks} [ICML’20]
\item CMVRL~\cite{hassani2020contrastive}: \textit{Contrastive Multi-View Representation Learning on Graphs} [ICML’20]
\item BGNN~\cite{zhu2020bilinear}: \textit{Bilinear Graph Neural Network with Neighbor Interactions} [IJCAI’20]
\item ADSF~\cite{zhang2020adaptive}: \textit{Adaptive Structural Fingerprints for Graph Attention Networks} [ICLR’20]
\end{itemize}

\vspace{+1mm}
\noindent
DAGNN~\cite{liu2020towards} and GCNII~\cite{chen2020simple} addressed the over-smoothing issue and presented very deep architectures (more than 50 layers).
CMVRL~\cite{hassani2020contrastive} introduced a self-supervised approach by contrasting structural views of graphs. 
BGNN~\cite{zhu2020bilinear} proposed a bilinear graph convolution operation which augments the weighted sum with pairwise interactions of neighboring nodes. 
ADSF~\cite{zhang2020adaptive} injected structural information into GAT, but, the range of the model is limited to 2-hop neighbors.
Table~\ref{res2020} summarizes the results.
We report the numbers from their papers. 
Even compared with very recent methods including deeper architectures such as DAGNN and GCNII, our CAD-Net still achieves state-of-the-art performance on CiteSeer and PubMed.

\begin{table*}[h]
\caption{Comparisons with the latest methods published in 2020 that share the same experimental settings as ours. We report the average accuracy (\%) where the numbers are from their papers. The best results are marked by bold. 
}
\label{res2020}
\centering
\footnotesize
\vspace{-0.5mm}
\begin{tabular}{lccc} 
    \toprule
    Method            
        &   {\tt\sc CiteSeer}      
        &   {\tt\sc Cora}     
        &   {\tt\sc PubMed} 
        \\
    \midrule
    DAGNN~\cite{liu2020towards}
        & $73.3$         
        & $84.4$         
        & $80.5$   
        \\
    GCNII~\cite{chen2020simple}
        & $73.2$ 
        & $85.3$ 
        & $80.3$ 
        \\
    CMVRL~\cite{hassani2020contrastive}
        & $73.3$ 
        & $\mathbf{86.8}$  
        & $80.1$  
        \\
    BGNN~\cite{zhu2020bilinear}
        & $74.0$ 
        & $84.2$ 
        & $79.8$   
        \\
    ADSF~\cite{zhang2020adaptive}
        & $74.0$  
        & $85.4$ 
        & $81.2$  
        \\
    \midrule
    CAD-Net 
        & $\mathbf{74.1}$ 
        & $84.3$  
        & $\mathbf{82.3}$  
        \\
    \bottomrule
\end{tabular}
\end{table*}

\subsection{Computational Complexity}
Note that the proposed AdaCAD does not require additional learning parameters since we utilize the classifier $g_{\phi}$.
Also, AdaCAD preserves the graph's sparsity and never needs to construct an $\mathbb{R}^{N \times N}$ matrix.
Instead, the second term in Eq. (\ref{eq:matrixform}) of the manuscript can be implemented by $K$ iterations of message passing as in~\cite{klicpera2018predict}.
Only one forward operation is additionally required to obtain our class-attentive transition probability.
To quantitatively measure the efficiency, we evaluated the average training time per epoch (ms) measured on a single Nvidia GTX 1080 Ti machine. 
Table~\ref{complexity} summarizes the results. 
As expected, CAD-Net is on par with APPNP and much faster than GAT, which validates the computational efficiency of the proposed method.
Note that GDC requires the inverse of $\mathbb{R}^{N \times N}$ matrix, which is computationally burden and raises singular error on Coauthor CS.

\begin{table*}[h]
\caption{Average training time per epoch (ms) measured on a single Nvidia GTX 1080 Ti machine.
For all experiments, we report the average time evaluated over 100 independent runs.
OOM denotes out-of-memory.
}
\vspace{-0.5mm}
\label{complexity}
\centering
\footnotesize
\begin{tabularx}{\textwidth}{cl*{7}{C}} 
    \toprule
    Type     &  Method   &  {\tt\sc CiteSeer}  &  {\tt\sc Cora}  &  {\tt\sc PubMed}  &  {\tt\sc Amazon Comp.}  &  {\tt\sc Amazon Photo}  &  {\tt\sc Coauthor CS}  &   {\tt\sc Coauthor Physics}\\
    \midrule
    \multirow{3}{*}{Vanilla}
    & Cheby   
        &   30.98     
        &   15.30     
        &   32.46 
        &   159.79       
        &   78.43 
        &   \scriptsize{(OOM)}   
        &   \scriptsize{(OOM)}
        \\
    & GCN     
        &   5.33     
        &   5.06     
        &   7.03  
        &   8.75       
        &   6.04
        &   12.13
        &   23.49
        \\
    & SAGE    
        &   3.90     
        &   3.81     
        &   3.92  
        &   7.33   
        &   4.68
        &   11.45
        &   22.17
        \\
    \midrule
    \multirow{3}{*}{\shortstack[1]{Extended \\ Aggregation}}
    & JK      
        &   5.82     
        &   5.55     
        &   6.84  
        &   14.05
        &   8.68
        &   14.36
        &   30.58
        \\
    & MixHop  
        &   21.09     
        &   14.42     
        &   24.40  
        &   98.95  
        &   50.79
        &   \scriptsize{(OOM)}  
        &   \scriptsize{(OOM)}
        \\
    & SGC     
        &   2.49     
        &   2.02     
        &   2.22  
        &   3.12
        &   2.51
        &   10.84  
        &   19.06
        \\
    \midrule
    \multirow{3}{*}{\shortstack[1]{Feature\\ Attention}}
    & AGNN    
        &   13.34     
        &   7.80     
        &   15.59  
        &   55.38   
        &   28.37
        &   26.09 
        &   62.67
        \\
    & GAT     
        &   9.40     
        &   9.21     
        &   18.37  
        &   108.19     
        &   46.94
        &   53.69  
        &   109.97 
        \\
    & U-Nets  
        &   30.86     
        &   33.51     
        &   49.90  
        &   48.57      
        &   37.21 
        &   68.53  
        &   210.05
        \\
    \midrule
    \multirow{2}{*}{\shortstack[1]{Graph\\ Diffusion}}
    & APPNP   
        &   7.16     
        &   6.84     
        &   8.03  
        &   16.05    
        &   8.36
        &   17.37
        &   32.22
        \\
    & GDC     
        &   49.13     
        &   50.66     
        &   579.81  
        &   741.46             
        &   119.80  
        &   \scriptsize{(OOM)}
        &   8776.93 
        \\
    \midrule
    Proposed
    & CAD-Net 
        &   6.55     
        &   7.60     
        &   8.63  
        &   21.98  
        &   12.79
        &   20.17             
        &   41.46
        \\
    \bottomrule
\end{tabularx}
\end{table*}

\subsection{Different Label Rates}
In this section we present the detailed experimental setup and results in tabular form for Figure~\ref{fig:label} of the manuscript.
From the given training set by the benchmark setting~\cite{yang2016revisiting} where 20 nodes per class are labeled, we randomly reduced the number of labeled nodes per class to 15, 10, and 5.
Table~\ref{res-label-citeseer}, \ref{res-label-cora}, and \ref{res-label-pubmed} show detailed results illustrated in Figure~\ref{fig:label} of the manuscript.

\subsection{Different Ranges}
In this section we present the detailed experimental setup and results in tabular form for Figure~\ref{fig:range} of the manuscript.
For Cheby, we used $K$-th order of Chebyshev polynomials to achieve the range of $K$.
For GCN, SAGE, JK, and GAT, we added additional layers before the last layer to achieve the desired range, whereas, for SGC, AGNN, APPNP, and the proposed CAD-Net, we simply increased the propagation steps.
Table~\ref{res-range-citeseer}, \ref{res-range-cora}, and \ref{res-range-pubmed} show the detailed results illustrated in Figure~\ref{fig:range} of the manuscript.

\clearpage
\begin{table*}
\caption{
Accuracy (\%) on {\tt\sc CiteSeer} with different numbers of labeled nodes per class.
The following tables show detailed results illustrated in Figure 2 of the manuscript.
We report the mean with the standard deviation (mean $\pm$ std) computed over 100 independent runs.
The best results are marked by bold. 
}
\label{res-label-citeseer}
\centering
\footnotesize
\begin{tabular}{lcccc} 
    \toprule
      &  \multicolumn{4}{c}{Labled nodes per class} \\
    \cmidrule(lr){2-5}
    Method        &  5  &  10  &  15  &  20   \\
    \midrule
    Cheby~\cite{defferrard2016convolutional}    & $58.4\pm1.3$      & $70.3\pm0.5$      & $70.9\pm0.6$      & $70.7\pm0.5$  \\
    GCN~\cite{kipf2016semi}                     & $57.1\pm2.5$      & $68.3\pm1.1$      & $71.1\pm0.8$      & $71.1\pm0.7$  \\
    SAGE~\cite{hamilton2017inductive}           & $60.1\pm1.8$      & $69.2\pm0.7$      & $70.9\pm0.6$      & $70.9\pm0.7$  \\
    JK~\cite{xu2018representation}              & $59.8\pm1.5$      & $63.6\pm2.0$      & $69.6\pm0.9$      & $69.2\pm0.9$  \\
    MixHop~\cite{abu2019mixhop}                 & $62.9\pm1.3$      & $69.1\pm1.1$      & $70.8\pm1.1$      & $71.5\pm0.8$  \\
    SGC~\cite{wu2019simplifying}                & $61.6\pm0.2$      & $68.5\pm0.2$      & $71.6 \pm 0.1$    & $71.9\pm0.1$  \\
    AGNN~\cite{thekumparampil2018attention}     & $62.0\pm2.0$      & $70.1\pm0.9$      & $71.3\pm0.8$      & $71.5\pm0.7$  \\
    GAT~\cite{velivckovic2017graph}             & $64.4\pm1.6$      & $70.1\pm0.8$      & $71.9\pm0.9$      & $72.5\pm0.8$  \\
    U-Nets~\cite{gao2019graph}                  & $ 60.8\pm1.4$     & $68.1\pm0.9$      & $ 69.9\pm0.7$     & $70.2\pm1.5$  \\
    APPNP~\cite{klicpera2018predict}            & $63.0\pm1.3$      & $70.5\pm0.7$      & $71.9\pm0.5$      & $71.8\pm0.5$  \\
    GDC~\cite{klicpera2019diffusion}            & $64.3\pm1.6$      & $70.1\pm0.9$      & $71.5\pm0.8$      & $72.7\pm0.8$  \\
    \midrule
    CAD-Net                 & $\mathbf{67.7\pm1.1}$ & $\mathbf{72.2\pm0.7}$ & $\mathbf{73.9\pm0.5}$ & $\mathbf{74.1\pm0.4}$ \\
    \bottomrule
\end{tabular}
\end{table*}

\begin{table*}
\caption{
Accuracy (\%) on {\tt\sc Cora} with different numbers of labeled nodes per class.
}
\label{res-label-cora}
\centering
\footnotesize
\begin{tabular}{lcccc} 
    \toprule
      &  \multicolumn{4}{c}{Labled nodes per class} \\
    \cmidrule(lr){2-5}
    Method        &  5  &  10  &  15  &  20   \\
    \midrule
    Cheby~\cite{defferrard2016convolutional}    & $59.9\pm2.5$      & $70.9\pm1.7$      & $79.6\pm0.6$      & $81.4\pm0.5$  \\
    GCN~\cite{kipf2016semi}                     & $69.6\pm1.6$      & $74.8\pm1.1$      & $80.1\pm0.9$      & $81.3\pm0.7$  \\
    SAGE~\cite{hamilton2017inductive}           & $71.0\pm1.6$      & $75.9\pm1.1$      & $80.2\pm0.7$      & $81.4\pm0.7$  \\
    JK~\cite{xu2018representation}              & $69.8\pm1.7$      & $73.5\pm1.3$      & $79.4\pm1.0$      & $81.2\pm0.8$  \\
    MixHop~\cite{abu2019mixhop}                 & $72.7\pm1.5$      & $76.2\pm1.2$      & $80.6\pm0.9$      & $82.0\pm0.8$  \\
    SGC~\cite{wu2019simplifying}                & $70.8\pm0.1$      & $74.7\pm0.1$      & $78.2\pm0.1$      & $81.0\pm0.2$  \\
    AGNN~\cite{thekumparampil2018attention}     & $74.0\pm1.2$      & $78.4\pm0.8$      & $82.1\pm0.8$      & $82.8\pm0.6$  \\
    GAT~\cite{velivckovic2017graph}             & $73.8\pm0.7$      & $78.6\pm0.6$      & $82.5\pm0.5$      & $83.1\pm0.8$  \\
    U-Nets~\cite{gao2019graph}                  & $73.6\pm1.8$      & $76.8\pm1.4$      & $81.4\pm0.9$      & $82.9\pm0.7$  \\
    APPNP~\cite{klicpera2018predict}            & $73.9\pm0.9$      & $77.5\pm0.8$      & $82.2\pm0.5$      & $82.9\pm0.5$  \\
    GDC~\cite{klicpera2019diffusion}            & $74.9\pm1.1$      & $78.6\pm0.9$      & $81.7\pm0.8$      & $82.7\pm0.7$  \\
    \midrule
    CAD-Net                 & $\mathbf{76.7\pm1.2}$ & $\mathbf{80.5\pm0.7}$ & $\mathbf{83.8\pm0.6}$ & $\mathbf{84.3\pm0.5}$ \\
    \bottomrule
\end{tabular}
\end{table*}

\begin{table*}
\caption{
Accuracy (\%) on {\tt\sc PubMed} with different numbers of labeled nodes per class.
}
\label{res-label-pubmed}
\centering
\footnotesize
\begin{tabular}{lcccc} 
    \toprule
      &  \multicolumn{4}{c}{Labled nodes per class} \\
    \cmidrule(lr){2-5}
    Method        &  5  &  10  &  15  &  20   \\
    \midrule
    Cheby~\cite{defferrard2016convolutional}    & $65.1\pm1.5$      & $71.1\pm1.6$      & $72.0\pm1.9$      & $75.2\pm1.4$  \\
    GCN~\cite{kipf2016semi}                     & $67.5\pm1.0$      & $75.2\pm0.6$      & $77.9\pm0.5$      & $79.0\pm0.5$  \\
    SAGE~\cite{hamilton2017inductive}           & $68.8\pm1.0$      & $74.8\pm0.7$      & $77.7\pm0.6$      & $78.7\pm0.4$  \\
    JK~\cite{xu2018representation}              & $70.3\pm1.2$      & $75.1\pm0.6$      & $78.1\pm0.4$      & $78.8\pm0.5$  \\
    MixHop~\cite{abu2019mixhop}                 & $69.1\pm2.0$      & $76.0\pm1.0$      & $78.8\pm0.6$      & $79.6\pm0.4$  \\
    SGC~\cite{wu2019simplifying}                & $66.1\pm0.5$      & $73.9\pm0.2$      & $76.7\pm0.1$      & $78.9\pm0.1$  \\
    AGNN~\cite{thekumparampil2018attention}     & $68.2\pm1.1$      & $74.4\pm0.7$      & $78.5\pm0.7$      & $79.3\pm0.8$  \\
    GAT~\cite{velivckovic2017graph}             & $67.3\pm1.0$      & $74.3\pm0.7$      & $78.5\pm0.4$      & $79.0\pm0.3$  \\
    U-Nets~\cite{gao2019graph}                  & $69.7\pm1.1$      & $74.6\pm0.7$      & $78.2\pm0.5$      & $77.9\pm0.9$  \\
    APPNP~\cite{klicpera2018predict}            & $69.4\pm0.9$      & $77.3\pm0.5$      & $79.3\pm0.5$      & $79.7\pm0.3$  \\
    GDC~\cite{klicpera2019diffusion}            & $68.2\pm1.0$      & $72.3\pm0.4$      & $77.5\pm0.4$      & $78.1\pm0.3$  \\
    \midrule
    CAD-Net                 & $\mathbf{77.1\pm1.4}$ & $\mathbf{78.9\pm1.5}$ & $\mathbf{81.5\pm0.4}$ & $\mathbf{82.3\pm0.4}$ \\
    \bottomrule
\end{tabular}
\end{table*}

\clearpage
\begin{table*}
\caption{
Accuracy (\%) on {\tt\sc CiteSeer} with varying ranges of the model.
The following tables show detailed results illustrated in Figure 3 of the manuscript.
We report the mean with the standard deviation (mean $\pm$ std) computed over 100 independent runs.
The best results are marked by bold. 
}
\label{res-range-citeseer}
\centering
\footnotesize
\begin{tabular}{lccccc}
\toprule
      &  \multicolumn{5}{c}{Range (hop)} \\
\cmidrule(lr){2-6}
Method              & 2         & 4         & 6         & 8         & 10        \\
\midrule
Cheby~\cite{defferrard2016convolutional}        & $70.7\pm0.5$      & $65.6\pm1.9$      & $60.8\pm4.1$      & $56.6\pm5.1$      & $56.8\pm3.9$      \\
GCN~\cite{kipf2016semi}                         & $71.1\pm0.7$      & $58.4\pm5.9$      & $35.0\pm9.6$      & $24.5\pm7.2$      & $21.3\pm2.3$      \\
SAGE~\cite{hamilton2017inductive}               & $70.9\pm0.7$      & $36.1\pm7.7$      & $19.1\pm2.5$      & $18.4\pm2.2$      & $19.2\pm2.7$      \\
JK~\cite{xu2018representation}                  & $69.1\pm1.1$      & $65.1\pm1.7$      & $63.8\pm1.8$      & $64.3\pm2.1$      & $63.6\pm2.4$      \\
SGC~\cite{wu2019simplifying}                    & $71.9\pm0.1$      & $70.8\pm0.1$      & $71.2\pm0.1$      & $71.2\pm0.3$      & $71.3\pm0.2$      \\
AGNN~\cite{thekumparampil2018attention}         & $70.9\pm0.9$      & $71.5\pm0.7$      & $71.4\pm0.8$      & $71.2\pm0.8$      & $71.0\pm0.7$      \\
GAT~\cite{velivckovic2017graph}                 & $72.5\pm0.8$      & $66.7\pm1.6$      & $62.0\pm4.4$      & $59.2\pm4.7$      & $51.2\pm10.2$     \\
APPNP~\cite{klicpera2018predict}                & $72.0\pm0.7$      & $71.8\pm0.5$      & $71.6\pm0.5$      & $71.7\pm0.5$      & $71.7\pm0.5$      \\
\midrule
CAD-Net            & $\mathbf{73.8\pm0.4}$      & $\mathbf{74.0\pm0.5}$      & $\mathbf{73.7\pm0.5}$      & $\mathbf{73.7\pm0.5}$      & $\mathbf{73.7\pm0.5}$      \\
\bottomrule
\end{tabular}
\end{table*}
\begin{table*}
\caption{
Accuracy (\%) on {\tt\sc Cora} with varying ranges of the model.
}
\label{res-range-cora}
\centering
\footnotesize
\begin{tabular}{lccccc}
\toprule
      &  \multicolumn{5}{c}{Range (hop)} \\
\cmidrule(lr){2-6}
Method         & 2    & 4    & 6    & 8    & 10   \\
\midrule
Cheby~\cite{defferrard2016convolutional}        & $81.4\pm0.5$      & $78.5\pm1.7$      & $71.7\pm5.7$      & $62.1\pm9.9$      & $57.1\pm10.4$  \\
GCN~\cite{kipf2016semi}                         & $81.3\pm0.7$      & $74.2\pm4.0$      & $49.0\pm15.7$     & $20.3\pm8.8$      & $20.8\pm8.6$  \\
SAGE~\cite{hamilton2017inductive}               & $81.4\pm0.7$      & $51.3\pm13.9$     & $20.5\pm9.1$      & $20.6\pm9.6$      & $22.1\pm10.1$  \\
JK~\cite{xu2018representation}                  & $81.2\pm0.8$      & $79.3\pm1.1$      & $78.7\pm1.7$      & $78.4\pm1.6$      & $78.3\pm2.2$  \\
SGC~\cite{wu2019simplifying}                    & $81.0\pm0.2$      & $81.0\pm0.3$      & $80.6\pm0.4$      & $80.4\pm0.4$      & $80.5\pm0.4$  \\
AGNN~\cite{thekumparampil2018attention}         & $82.8\pm0.6$      & $81.7\pm0.7$      & $81.5\pm0.7$      & $81.4\pm0.6$      & $81.3\pm0.5$  \\
GAT~\cite{velivckovic2017graph}        & $\mathbf{83.1\pm0.8}$      & $80.5\pm1.3$      & $76.6\pm2.5$      & $75.4\pm2.4$      & $74.8\pm2.8$  \\
APPNP~\cite{klicpera2018predict}                & $81.2\pm0.6$      & $81.2\pm0.6$      & $82.2\pm0.8$      & $82.2\pm0.9$      & $82.0\pm0.9$  \\
\midrule
CAD-Net             & $82.5\pm0.5$      & $\mathbf{83.8\pm0.4}$    & $\mathbf{84.3\pm0.5}$      & $\mathbf{84.2\pm0.4}$      & $\mathbf{83.8\pm0.5}$ \\
\bottomrule
\end{tabular}
\end{table*}
\begin{table*}
\caption{
Accuracy (\%) on {\tt\sc PubMed} with varying ranges of the model.
}
\label{res-range-pubmed}
\centering
\footnotesize
\begin{tabular}{lccccc}
\toprule
      &  \multicolumn{5}{c}{Range (hop)} \\
\cmidrule(lr){2-6}
Method         & 2    & 4    & 6    & 8    & 10   \\
\midrule
Cheby~\cite{defferrard2016convolutional}        & $75.2\pm1.4$      & $72.6\pm3.4$      & $69.8\pm3.4$      & $70.0\pm4.7$      & $71.5\pm5.3$  \\
GCN~\cite{kipf2016semi}                         & $79.0\pm0.5$      & $75.0\pm3.9$      & $65.0\pm9.9$      & $43.7\pm2.8$      & $39.4\pm1.0$  \\
SAGE~\cite{hamilton2017inductive}               & $78.7\pm0.4$      & $66.4\pm9.2$      & $41.8\pm3.4$      & $41.0\pm0.3$      & $41.0\pm0.3$  \\
JK~\cite{xu2018representation}                  & $78.7\pm0.5$      & $75.8\pm1.8$      & $74.7\pm2.6$      & $74.9\pm2.3$      & $75.3\pm2.1$  \\
SGC~\cite{wu2019simplifying}                    & $78.9\pm0.1$      & $79.4\pm0.1$      & $77.7\pm0.2$      & $77.1\pm0.4$      & $75.9\pm0.3$  \\
AGNN~\cite{thekumparampil2018attention}         & $78.9\pm0.6$      & $79.3\pm0.8$      & $77.9\pm0.6$      & $77.2\pm0.6$      & $76.7\pm0.5$  \\
GAT~\cite{velivckovic2017graph}                 & $79.0\pm0.3$      & $74.6\pm4.1$      & $73.1\pm4.6$      & $73.3\pm6.0$      & $73.0\pm3.3$  \\
APPNP~\cite{thekumparampil2018attention}       & $\mathbf{79.3\pm0.3}$      & $79.7\pm0.3$      & $80.1\pm0.3$      & $80.1\pm0.2$      & $80.2\pm0.3$  \\
\midrule
CAD-Net             & $79.2\pm0.7$      & $\mathbf{81.3\pm0.5}$      & $\mathbf{82.0\pm0.4}$      & $\mathbf{82.3\pm0.4}$      & $\mathbf{81.9\pm0.3}$ \\
\bottomrule
\end{tabular}
\end{table*}

\end{document}